\documentclass{ceurart}

\usepackage{amsmath}
\usepackage{amssymb}
\usepackage{libertinust1math}
\usepackage{booktabs}
\usepackage{subcaption}
\usepackage{tikz}
\usetikzlibrary{arrows.meta, positioning, fit, backgrounds, calc}

\newcommand{\code}[1]{\texttt{#1}}

\RenewDocumentCommand\nonumnote{m}{}
\RenewDocumentCommand\logonote{m}{}

\begin{document}

\title{A GPU-Resident Pipeline for Multi-Robot 3D Inspection with Limited Perception}

\author[1]{Kry\v{s}tof P\v{s}eni\v{c}ka}[%
email=krystofpsenicka13@gmail.com,
]
\author[1]{Stefan Edelkamp}
\address[1]{Department of Theoretical Computer Science and Mathematical Logic,
  Faculty of Mathematics and Physics, Charles University, Prague, Czech Republic}

\begin{abstract}
We present a complete, GPU-resident pipeline for multi-robot 3D inspection of
underwater structures. Given a triangle mesh and a fleet of $K$ robots, each
carrying a short-range, limited field-of-view camera, the pipeline computes
6-DOF trajectories that jointly cover a target fraction of the surface while
minimizing a linearly interpolated makespan/total-cost objective. The joint
problem is NP-hard, so we decompose it into five core stages --- candidate
viewpoint generation, visibility computation, set cover, min-max vehicle routing
solved as a mixed integer program (MIP) with lifted Miller--Tucker--Zemlin
subtour elimination, and priority-based multi-agent path finding with
continuous-time inter-robot conflict checking --- flanked by mesh loading and
surface sampling as preprocessing and by trajectory smoothing as
postprocessing. Our contributions are: an end-to-end GPU-accelerated
integration of the five stages, with its kernel design documented; a GPU batch
adaptation of Lien's $\varepsilon$-visibility with four custom CUDA kernels; a $\beta$-aware per-vehicle tour upper bound derived from
the interpolated objective, supplementing the Desrochers--Laporte coefficient
lift with forbidden-pair cuts; a 4D space-time extension of Zhou and Zeng's GPU
parallel-frontier A*; and an experimental evaluation on 14 meshes (up to
$7.2$M faces) that ablates the routing cuts and the parallel-frontier search,
audits collision freedom and waypoint attainment \emph{after} trajectory
smoothing, and measures the end-to-end gap against analytical and LP-relaxation
lower bounds. Code and raw results are
publicly available.
\end{abstract}


\maketitle

\section{Introduction}
\label{sec:intro}

Autonomously inspecting the surface of a complex three-dimensional structure
underwater is important across engineering and science: commercial shipping
relies on regular hull inspection for biofouling and corrosion; offshore
wind-farm operators need periodic condition assessment of monopile foundations;
harbor and dam operators inspect submerged walls; and underwater archaeology and
marine biology require dense photographic surveys of wrecks and coral
reefs~\cite{inspectunderwater,carreno2020,ThompsonG19}. These tasks share a
common structure: a fixed surface must be visually covered, and each surface
region has to be seen from within a limited range by a camera with a finite
field of view. Single-robot solutions scale poorly; a fleet of small autonomous
underwater vehicles (AUVs) can cover a complex surface faster and can cover for
each other if one vehicle fails mid-mission.

The underwater setting motivates two design choices. Perception range is short,
so the planner must bring viewpoints close to the surface (our camera far plane
defaults to $6\,\text{m}$). And because in-mission re-tasking and high-bandwidth
communication are limited, careful \emph{pre-mission planning} is the dominant
lever for mission success, so we target offline planning.

\paragraph{Scope and assumptions.}
We address the \emph{offline planning layer} in isolation, under a deliberately
idealized model of everything around it: a geometric prior exists as a mesh
(from a bathymetric or photogrammetric survey), localization, map registration
and execution are exact, and visibility is a deterministic function of geometry.
Currents, acoustic bandwidth, energy, sensor noise and turbidity are not
modelled --- they inform our defaults (short far plane, low cruise speed,
generous margins) but do not enter the
optimization. The evaluation is therefore deterministic simulation, and every
reported quantity is a property of the \emph{plan}, not of a deployment. These
assumptions make the joint problem separable and let us isolate each stage's
contribution; Section~\ref{sec:conclusion} says which a real mission would
violate first.

We describe \textsc{3D-Inspection}, a framework that takes a triangle mesh, a
fleet size $K$ and a coverage threshold, and outputs a collision-free 6-DOF
trajectory per robot whose union of seen surface exceeds the threshold, while
minimizing a user-weighted interpolation of makespan and total cost. The joint
formulation is NP-hard --- it combines the 3D watchman route, vehicle routing
and multi-agent path finding --- so the framework is a \emph{decomposition} that
solves each subproblem with strong approximations, leaning on GPU acceleration
to keep end-to-end runtimes in the tens of minutes at $|P|\sim10^5$.

\paragraph{Contributions.}
Most individual stages instantiate established methods; we state explicitly
which parts are new algorithmic work and which are systems integration.
\emph{Algorithmic:}
\begin{enumerate}
\item A GPU batch adaptation of Lien's $\varepsilon$-visibility~\cite{lien2007}
as four custom CUDA kernels over a compacted (viewpoint, point) pair list,
replacing the divergent per-viewpoint radial partitioning of a naive port with
one sort plus two atomic scatter passes (Section~\ref{sec:gpu}), evaluated
against bounding-volume-hierarchy (BVH) raycasting as ground truth.
\item A min-max vehicle routing MIP combining the lifted Miller--Tucker--Zemlin
subtour elimination of Desrochers and Laporte~\cite{desrochers1991} with two
original additions --- a $\beta$-aware per-vehicle tour upper bound derived from
the interpolated objective, used to tighten arc-filtering cuts, and
forbidden-pair cuts for jointly infeasible arc pairs (Section~\ref{sec:vrp});
both are ablated in Section~\ref{sec:experiments}.
\item A 4D $(t,x,y,z)$ space-time extension of Zhou and Zeng's GPU
parallel-frontier A*~\cite{zhouzeng2015} whose successor filter is a closed-form
continuous-time separation test against earlier robots' committed motion rather
than a discrete cell reservation (Section~\ref{sec:mapf}), forming the
single-agent core of a priority-based planner after Silver~\cite{silver2005};
ablated in Section~\ref{sec:experiments}, where the parallel frontier is worth
a median $38\times$ at no measured cost in path quality.
\end{enumerate}
\emph{Integration and evaluation:} an end-to-end GPU-resident pipeline with a
documented mapping of stages onto kernels, and a comparative evaluation of
sampling strategies, visibility models, set-cover heuristics, the routing cuts
and the multi-agent layer against LP-relaxation and analytical lower bounds,
including an audit of collision freedom and waypoint attainment on the final
smoothed trajectories.

\section{Problem Statement}
\label{sec:problem}

A discrete triangle mesh $\mathcal{M} = (V_{\text{mesh}},F)$ describes the
inspected structure, where
$V_{\text{mesh}} = \{\mathbf{v}_1,\ldots,\mathbf{v}_{|V_{\text{mesh}}|}\}
\subset\mathbb{R}^3$ is the vertex set and
$F \subseteq V_{\text{mesh}}^3$ the set of triangular faces, each an ordered
vertex triple whose winding fixes an outward orientation. We densely sample
$\mathcal{M}$ to obtain a finite set
$P = \{\mathbf{p}_1,\ldots,\mathbf{p}_M\}$ of \emph{target points}
$\mathbf{p}_i \in \mathbb{R}^3$, together with an associated outward unit normal
$\mathbf{n}_i \in \mathbb{R}^3$ ($\lVert\mathbf{n}_i\rVert = 1$) at each
$\mathbf{p}_i$. A robot \emph{pose} is a pair $(\mathbf{x},R)$, with
$\mathbf{x}\in\mathbb{R}^3$ the camera position and $R\in SO(3)$ its
orientation: the camera looks along the first column of $R$, the other two
spanning the image plane. Coverage is measured on $P$: a point $\mathbf{p}_i$ is
covered by a trajectory $\pi$ when some pose $(\mathbf{x},R) \in \pi$ sees it
through the robot's camera (a pyramid frustum with specified field of view,
aspect and near/far range) without occlusion by the mesh. The planner works with
a finite \emph{candidate viewpoint set} $C = \{c_1,\ldots,c_N\}$, each
$c_j = (\mathbf{x}_j,R_j)$ a sampled free-space pose, from which a covering
subset is selected (Section~\ref{sec:pipeline}).

Let $\mathcal{X}_{\text{free}} \subset \mathbb{R}^3$ denote the free
configuration space (positions whose robot center does not collide with the
inflated mesh). A \emph{trajectory}
$\pi_r\colon [0,T_r] \to \mathcal{X}_{\text{free}}\times SO(3)$ for robot $r$ is
a continuous 6-DOF path from the robot's home pose. Writing
$\mathrm{Vis}(\pi_r,\mathcal{M})\subseteq P$ for the target points visible from
any pose along $\pi_r$, the joint coverage of a fleet with trajectories
$\pi_1,\ldots,\pi_K$ is
$\mathrm{Coverage}(\pi_1,\ldots,\pi_K) =
\frac{1}{|P|}\bigl|\bigcup_{r=1}^{K}\mathrm{Vis}(\pi_r,\mathcal{M})\bigr|$.
Given a target coverage $\alpha \in (0,1]$, an objective weight
$\beta \in [0,1]$, and a safety distance $d_{\text{safe}}$, the joint planning
problem is
\begin{equation}\label{eq:joint-problem}
\begin{aligned}
\min_{\pi_1,\ldots,\pi_K}\;\; & \beta\cdot \max_{r} T_r + (1-\beta)\cdot \sum_{r=1}^{K} T_r\\
\text{s.t.}\;\; & \mathrm{Coverage}(\pi_1,\ldots,\pi_K) \geq \alpha, \\
& \pi_r(0) = (\mathbf{x}_r^0,R_r^0) \quad \forall\, r,\\
& \pi_r(t)\in \mathcal{X}_{\text{free}}\times SO(3) \quad \forall\, r,t,\\
& \|\mathbf{x}_r(t) - \mathbf{x}_s(t)\| \geq d_{\text{safe}} \quad \forall\, r\neq s,\forall\, t.
\end{aligned}
\end{equation}
The objective interpolates the two natural fleet-routing criteria,
\emph{makespan} (when the last robot finishes) and \emph{total cost} (the sum of
per-robot durations); the last constraint enforces pairwise separation at all
times.

\paragraph{Hardness.}
Problem~\eqref{eq:joint-problem} is NP-hard. For a single robot it reduces to
the 3D \emph{watchman route} problem~\cite{watchman}, a generalization of the 2D
art gallery problem. Even with a fixed viewpoint set, selecting a minimum subset
covering $\alpha |P|$ points is set cover, NP-hard and inapproximable beyond a
logarithmic factor~\cite{chvatal1979,vazirani2001}; routing the selected
viewpoints among vehicles is a VRP; and the time-consistent collision constraint
is multi-agent path finding (MAPF), NP-hard even on a discrete
grid~\cite{sharon2015}. We therefore pursue a decomposition.

\section{Related Work}
\label{sec:related}

\paragraph{Guarding and coverage.}
Classical 2D \emph{art gallery} problems ask for a minimum set of guards seeing
every interior point of a polygon; the minimum is NP-hard and APX-hard, both of
which carry over to 3D~\cite{leelin1986,eidenbenz2001}. The \emph{watchman
route}~\cite{watchman} lets the guard move, and hitting-set formulations reduce
3D surface inspection to a discrete geometric set cover, the approach we follow.
Coverage path planning~\cite{galceran2013} and next-best-view methods address
related problems; Bircher et al.~\cite{bircher2015} in particular iterate
set-cover-style viewpoint selection with a TSP-style tour, mirroring our
two-stage structure.

\paragraph{Inspection planning in robotics.}
The robotics literature attacks the same task from the opposite end, coupling
planning with perception rather than assuming it away. Englot and
Hover~\cite{englot2013} plan 3D ship-hull coverage by sampling feasible views on
a bathymetric prior, refine the tour with a sampling-based redundant roadmap and
close the loop against the map the vehicle actually builds, with sensing
uncertainty an explicit planning term; Hern\'andez et al.~\cite{hernandez2015}
plan online in unknown underwater environments; McMahon et
al.~\cite{inspectunderwater} interleave survey and inspection so the map
improves mid-mission. Our formulation is by contrast strictly \emph{open loop}.
That buys global structure --- a min-max fleet objective over the whole surface,
and inter-robot coordination an incremental next-best-view planner cannot
express --- at the cost of robustness to map error; a practical system would use
a plan like ours as the mission skeleton and an active-perception layer to
repair it. Note also that many deployed systems use tethered ROVs, where tether
management and operator workflow rather than routing dominate mission design; we
target untethered survey-class AUV fleets.

\paragraph{Visibility.}
A practical visibility model answers two questions: is a point in the frustum,
and is it unoccluded? A fast pyramid-frustum test~\cite{pyramidvis} with
back-face culling filters candidates; the rest are confirmed by ray-cast
occlusion, accelerated by a bounding-volume hierarchy
(BVH)~\cite{rubinwhitted1980} to $O(\log|F|)$ per ray --- we use the GPU BVH
builder and traversal of NVIDIA OptiX~\cite{optix}. Lien~\cite{lien2007}
introduces \emph{$\varepsilon$-visibility}, an approximate model on oriented
point sets that never queries a mesh, so it still applies when the input is only
a point cloud. We reuse Lien's test (adapted to the GPU) as a scoring primitive
but not his outer star-shaped decomposition loop.

\paragraph{Set cover.}
The \emph{greedy} algorithm achieves an $H(d)$ approximation
ratio~\cite{chvatal1979}, essentially tight~\cite{vazirani2001}. Minoux's
\emph{Accelerated Greedy}~\cite{minoux1978} exploits submodularity: marginal
gains are nonincreasing, so a stale gain is a valid upper bound and most
candidates need never be re-scored --- identical selections to plain greedy,
orders of magnitude faster. The LP relaxation of the set-cover ILP gives the
lower bound against which we measure the optimality gap.

\paragraph{Vehicle routing.}
The VRP generalizes the TSP to a fleet, each vehicle starting and ending at a
depot. Subtours are eliminated either by the exponential
Dantzig--Fulkerson--Johnson family~\cite{dantzig1954} or the polynomial-size
Miller--Tucker--Zemlin (MTZ) family~\cite{miller1960}; Desrochers and
Laporte~\cite{desrochers1991} strengthen MTZ into a tighter \emph{lifted MTZ},
whose LP bound is $15$--$25\%$ stronger than the assignment relaxation on
symmetric random distance instances --- exactly our regime, since our distance
matrix comes from shortest paths on a voxel graph, computed with NVIDIA
cuGraph~\cite{cugraph} rather than Euclidean distance, which would ignore
obstacles.

\paragraph{Multi-agent path finding.}
Silver's cooperative pathfinding~\cite{silver2005} fixes a priority ordering and
plans agents one at a time, reserving each path in a space-time table: fast and
linear-scaling but incomplete in general, with ordering sensitivity mitigated by
trying several orderings. Conflict-Based Search (CBS)~\cite{sharon2015} is
optimal and complete at worst-case exponential cost. The low-level search is
space-time A*; Zhou and Zeng~\cite{zhouzeng2015} parallelize A* on a GPU by
expanding whole frontier layers at once. A discrete vertex/edge reservation
guarantees conflict freedom only on the grid, not for the continuous motion
actually executed; continuous-time MAPF~\cite{ccbs} instead tests the
closed-form distance between two moving bodies, the route we take at the low
level. Priority-based planning suits sparse conflicts (small fleets, large
environments), CBS dense ones; our setting is the former.

\section{Pipeline}
\label{sec:pipeline}

\begin{figure}[t]
\centering
\resizebox{\linewidth}{!}{%
\begin{tikzpicture}[
  stage/.style={
    rectangle, draw=black!70, rounded corners=2pt,
    minimum width=15mm, minimum height=11mm, text width=15mm, align=center,
    inner sep=1pt, font=\scriptsize, text=black,
  },
  core/.style={stage, fill=gray!15},
  pre/.style={stage, fill=white, dash pattern=on 3pt off 2pt, draw=black!55, text=black!75},
  arr/.style={-{Stealth[length=4pt]}, thick, black!70},
  io/.style={font=\scriptsize\itshape, text=black!60, text width=13mm, align=center},
  node distance=3mm,
]
\node[io] (in) {mesh, $K$, $\alpha$, $\beta$};
\node[pre, right=of in] (pre1) {load \& sample};
\node[core, right=of pre1] (c1) {\textbf{1}\\viewpoints};
\node[core, right=of c1] (c2) {\textbf{2}\\visibility};
\node[core, right=of c2] (c3) {\textbf{3}\\set cover};
\node[core, right=of c3] (c4) {\textbf{4}\\routing};
\node[core, right=of c4] (c5) {\textbf{5}\\MAPF};
\node[pre, right=of c5] (post1) {smooth \& densify};
\node[io, right=of post1] (out) {6-DOF traj.};
\draw[arr] (in) -- (pre1);
\draw[arr] (pre1) -- (c1);
\draw[arr] (c1) -- (c2);
\draw[arr] (c2) -- (c3);
\draw[arr] (c3) -- (c4);
\draw[arr] (c4) -- (c5);
\draw[arr] (c5) -- (post1);
\draw[arr] (post1) -- (out);
\end{tikzpicture}}
\caption{The pipeline: five core stages (shaded) with thin pre- and
postprocessing (dashed).}
\label{fig:pipeline}
\end{figure}
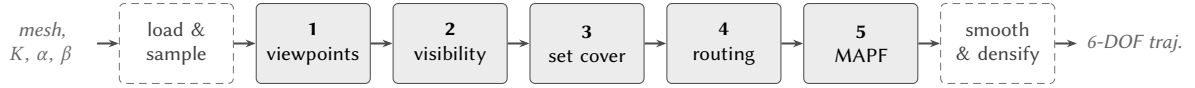

We organize the planner as a \emph{five-stage decomposition} of the joint
problem, flanked by thin preprocessing (mesh loading, surface sampling) and
postprocessing (coarse routes to dense trajectories) phases;
Figure~\ref{fig:pipeline}. Each stage maps one-to-one onto a canonical
subproblem from the hardness sketch and is efficiently solvable with existing
approximations. The cost is global optimality: chosen viewpoints do not
anticipate travel cost, and routes do not anticipate path-finding conflicts. We
quantify the resulting gap against lower bounds in
Section~\ref{sec:experiments}.

\subsection{Preprocessing and Viewpoint Sampling}

The mesh is read (GLB/OBJ/STL via \code{trimesh}), rescaled so its longest axis
matches a target length, and Poisson-disk-sampled to obtain $P$ with outward
normals. Two voxel structures are built: an occupancy grid inflated by
$r_{\text{robot}}$, and a Euclidean signed distance field (ESDF). A
\emph{feasible sampling region} keeps free voxels whose signed distance lies in
$[2r_{\text{robot}},\,0.95 z_f]$ (cameras neither too close nor beyond the far
plane), each weighted by $\mathrm{SDF}(\mathbf{x})^2$ to bias toward a large
frustum footprint.

The default \code{WeightedViewpointSampler} draws positions from the cumulative distribution of this
weight grid, orients each toward the centroid of its nearest target points, and
adds bounded angular noise; optional \emph{curvature weighting} multiplies the
weight by a $k$-nearest-neighbor mean-normal-deviation factor, biasing candidates toward locally
complex regions. Two iterative variants exist: a \emph{targeted} resampler
(following Glorieux et al.~\cite{glorieux2020}) biased toward under-covered
regions, and a covariance matrix adaptation evolution strategy (\emph{CMA-ES}) sampler treating each new 6-DOF pose as a
continuous optimization of deficit-weighted coverage gain minus travel cost.

\subsection{Visibility}
\label{sec:vis}

This stage computes $V \in \{0,1\}^{N\times M}$ with $V_{ji}=1$ iff target
$\mathbf{p}_i$ is visible from candidate $c_j$. A shared GPU frustum-culling
kernel launches one thread per (viewpoint, point) pair, doing the three FOV
inequality tests and a back-face check in registers; one of two occlusion models
then confirms survivors. \emph{BVH raycasting} (default) uses the Triro binding
of NVIDIA OptiX: the BVH is built once from the mesh triangles and all surviving
rays are submitted in one batch, a nearest hit strictly closer than the target
meaning occlusion. \emph{$\varepsilon$-visibility} operates directly on $P$ and
its normals without consulting the mesh, so it survives point-cloud-only input:
the angular bin width $\varepsilon$ follows from the local sampling density and
a per-viewpoint characteristic distance, frustum survivors are binned into a
latitude--longitude grid, the minimum back-facing distance per bin is recorded
by atomic-min, and a front-facing point is visible iff no back-facing point in
its bin is closer. Section~\ref{sec:gpu} details the four batch CUDA kernels.

\subsection{Set Cover}

We select the smallest $S^\star\subseteq C$ whose union visibility exceeds
$\alpha|P|$. The default \code{LazyGreedySetCover} implements Minoux's
accelerated greedy: a max-heap keyed on cached marginal-gain upper bounds with a
generation counter, re-evaluating a popped candidate only if stale and never
re-scoring one whose stale bound is below the best fresh gain. A GPU variant
instead scores all candidates as one $V\mathbf{u}$ matrix--vector product (GEMV). An optional
\emph{expansion} refinement re-samples each selected pose within a small sphere
to reduce the count further.

\subsection{Vehicle Routing}
\label{sec:vrp}

Selected viewpoints become robot-center waypoints and the $K$ home positions are
appended as fixed depots. The pairwise distance matrix is computed as shortest
paths on a 26-connected free-space voxel graph via cuGraph's GPU-parallel single-source shortest paths.
We then solve a min-max VRP mixed integer program (MIP; HiGHS on CPU or cuOpt on GPU) with binary arc
variables $x_{ij}^v$, MTZ rank variables $u_i$ and a makespan variable $T$,
interpolating makespan and total distance:
\begin{equation}
\min\quad \beta\cdot \frac{T}{T_{\text{norm}}} + (1-\beta)\cdot \frac{\sum_{v,i,j} c_{ij}\, x_{ij}^v}{C_{\text{norm}}}.
\end{equation}
Both $T$ and $\sum c_{ij}x_{ij}^v$ are \emph{distances} in metres ($c_{ij}$ is a
voxel-graph shortest path); they become times only after MAPF. The two terms
have very different magnitudes --- one tour versus $K$ tours --- so each is
scaled by a constant derived from the instance. With
$T_{\text{lb}} = \max_{j}\min_{v}(c_{d_v,j}+c_{j,d_v})$, the round trip forced
by the hardest customer and a valid makespan lower bound, we set
$T_{\text{norm}} = T_{\text{lb}}$ and $C_{\text{norm}} = K\,T_{\text{lb}}$, so
both terms are $O(1)$ and $\beta=0.5$ genuinely weights the two criteria equally
rather than being dominated by whichever is numerically larger.
Subtours are removed by the lifted MTZ inequality of~\cite{desrochers1991},
$u_i - u_j + n_c x_{ij}^v + (n_c-2)x_{ji}^v \le n_c-1$, plus a
multi-depot-adapted lifted upper bound
$u_i \le n_c - (n_c-1)\sum_v x_{d_v,i}^v$ pinning a customer visited directly
from its depot to position~1. We omit the symmetric lifted lower bound, valid
only for single-TSP tours of length exactly $n_c$, which would cut off feasible
multi-depot solutions.

\paragraph{$\beta$-aware arc filtering.}
Two original cuts tighten the model. From the nearest-neighbor warm-start's
objective we derive a per-tour upper bound valid for every feasible optimum,
\begin{equation}
T_{\text{tour}}^{\text{ub}} =
\begin{cases}
T_{\text{ws}} & \beta = 1, \\
C_{\text{ws}} & \beta = 0, \\
\min\!\bigl( T_{\text{ws}} + \tfrac{1-\beta}{\beta} C_{\text{ws}}
             \tfrac{T_{\text{norm}}}{C_{\text{norm}}},\, C_{\text{ws}} \bigr)
  & 0 < \beta < 1,
\end{cases}
\end{equation}
where $T_{\text{ws}}, C_{\text{ws}}$ are the warm-start makespan and total cost.
Since the distance matrix satisfies the triangle inequality, any arc $(i,j)$ on
vehicle $v$ with $c_{d_v,i}+c_{ij}+c_{j,d_v} > T_{\text{tour}}^{\text{ub}}$ can
be dropped (reachability filter, adapting~\cite{desrochers1991} Prop.~6); the
same argument on two consecutive arcs yields \emph{forbidden-pair cuts}
$x_{ki}^v + x_{ij}^v \le 1$ for predecessors $k$ that would drive the minimum
tour cost above the bound. Both back-ends are warm-started by a nearest-neighbor
heuristic.

\subsection{Multi-Agent Path Finding and Postprocessing}
\label{sec:mapf}

The VRP assigns each robot a sequence of waypoints; MAPF computes the space-time
paths connecting them. We plan robots one at a time in priority order, committing
each finished trajectory as a \emph{committed motion} --- its coarse world
polyline plus timing --- against which every later robot is checked. Up to 20
orderings (longest-first, shortest-first, random) are tried and the best
conflict-free $\beta$-blended plan kept. We choose priority-based planning over
CBS because at our scale ($K=2$--$6$) conflicts are rare, the VRP having already
spread robots across largely disjoint regions, and because it suits our
GPU-first single-agent search.

\paragraph{Continuous-time conflicts.}
A discrete space-time reservation table rules out \emph{vertex} conflicts only.
The executed trajectory interpolates between cell centres, so two robots can
still swap cells, cross diagonally, or dwell in a cell a passing robot sweeps
through, without ever sharing a cell at the same index. We therefore replace the
reservation lookup by an exact test on the interpolated motion, after
continuous-time MAPF~\cite{ccbs}: a candidate move
$\mathbf{a}_1\!\to\!\mathbf{b}_1$ over one step is rejected if, for any
committed robot moving $\mathbf{a}_2\!\to\!\mathbf{b}_2$ over an overlapping
interval,
$\min_{t\in[0,1]}\lVert(\mathbf{a}_1+t\Delta_1)-(\mathbf{a}_2+t\Delta_2)\rVert$
with $\Delta_i = \mathbf{b}_i-\mathbf{a}_i$ falls below
$d_{\text{safe}}=2r_{\text{robot}}$. That minimum is closed form (clamp a
quadratic's stationary point to $[0,1]$), so the test is a few fused
multiply--adds per (candidate, segment) pair and vectorizes over the successor
batch. Since the dwell at an inspection pose must also be safe, a goal cell is
accepted only if \emph{reachable and holdable}: the robot must stay clear for
its whole dwell plus any later steps in the same voxel. Edge and swap conflicts
are thus excluded by construction.

\paragraph{Single-agent search.}
The single-agent core is a CuPy port of Zhou and Zeng's heuristic-guided
parallel-frontier A*~\cite{zhouzeng2015}, retargeted from pure spatial search to
a 4D $(t,x,y,z)$ state in the sense of Silver~\cite{silver2005}. Each state has
27 successors: the 26 grid neighbors, each costing $w\,\Delta + \Delta_t$ with
$w\in\{1,\sqrt2,\sqrt3\}$ the neighbor's length in cells, $\Delta$ the cell size
and $\Delta_t$ a per-step time charge, plus a wait costing $\Delta_t$. The
heuristic is
\begin{equation}
h(t,x,y,z,\mathbf{g}) = \Delta\lVert(x,y,z)-\mathbf{g}\rVert_2 + \Delta_t
\lVert(x,y,z)-\mathbf{g}\rVert_\infty .
\end{equation}
It is admissible because the two cost components are additive and each term
underestimates its own: any path to $\mathbf{g}$ is a grid polyline, so its
spatial cost $\sum_k w_k\Delta$ is at least $\Delta\lVert\cdot\rVert_2$; and
since one step changes each coordinate by at most one cell, at least
$\lVert\cdot\rVert_\infty$ steps are needed, each charging $\Delta_t$. Waits and
detours only add cost and the inter-robot filter only removes successors, so
neither exposes a cheaper path. The parallel expansion is nevertheless the usual
bounded relaxation: expanding all frontier cells with $f\le f_{\min}+f_\Delta$
trades optimality for occupancy, so paths are $f_\Delta$-suboptimal --- a worst
case that E17 finds is never paid at our default $f_\Delta=2.0$.

\paragraph{Postprocessing.}
Each leg is shortcut and B-spline-smoothed with the Open Motion Planning Library (OMPL), assigned yaw and camera
pitch consistent with the inspection poses, and densified to
$\Delta_{\text{traj}}=0.02\,\text{s}$ for Isaac~Sim replay. Smoothing runs
\emph{after} the conflict test, so it could move a path off the motion it was
cleared against. Environment clearance survives by construction, as the OMPL
shortcut validates against the fine occupancy grid inflated by
$r_{\text{robot}}$; inter-robot safety is restored by re-running the same
closed-form segment test on the resampled smoothed leg, falling back to the
unsmoothed A* polyline if any segment fails. Section~\ref{sec:experiments}
audits the final densified output directly rather than relying on this argument.

\subsection{GPU Implementation}
\label{sec:gpu}

Every stage except the sequential priority loop of MAPF, the MIP control flow
and mesh I/O is device-resident, and the large arrays --- $V$, points and
normals, occupancy grid and ESDF, $g$-cost and predecessor volumes --- are
allocated once and never round-trip to the host. Frustum culling launches one
thread per (viewpoint, point) pair over an $N\times M$ grid, doing three dot
products and a back-face test in registers and writing one byte; point-array
reads are warp-coalesced and the viewpoint warp-uniform. Survivors are compacted
into a pair list, so every later kernel is one-dimensional over $R\ll NM$ pairs.
The four
$\varepsilon$-kernels run over that list: \code{batch\_epsilon\_preprocess}
(per-pair distance, spherical angles, front/back-facing flag),
\code{pair\_minmax} (per-viewpoint angular extent by atomics),
\code{batch\_scatter\_min\_occluders} (back-facing distances scattered into
per-viewpoint latitude--longitude bins in one flat buffer addressed by a
CSR-style \code{bin\_offsets} prefix sum, via \code{atomicMin} on the float bit
pattern), and \code{batch\_check\_visibility}. The key deviation from a direct
port of Lien's algorithm is that points are sorted by radial bin \emph{before}
the scatter, turning the original's data-dependent per-viewpoint partitioning
loop --- which serializes badly inside a warp --- into a contiguous streaming
write; this is what keeps the batch form within ${\sim}2\times$ of a hardware
BVH trace rather than orders of magnitude behind (Table~\ref{tab:e05}). Set
cover on GPU is one memory-bound $V\mathbf{u}$ GEMV of $NM$ bytes per iteration,
which loses to a host-side lazy-greedy heap at $|C|=1500$ but wins once $|C|$
reaches the tens of thousands. In space-time A*, the $E\times27$ successor batch
is filtered by vectorized masks and duplicates are resolved without atomics by
one lexicographic sort keyed on (target cell, tentative $g'$), so the first
entry of each run is the best parent --- a deterministic, warp-friendly
alternative to per-cell locking. The remaining bottleneck is the
$O(T\!\times\!N_xN_yN_z)$ $g$-cost volume, allocated only over the current leg's
local time horizon, which is what keeps a $50\,\text{m}$ Duke instance inside
$24\,\text{GB}$.

\section{Experimental Evaluation}
\label{sec:experiments}

\paragraph{Setup.}
Experiments run on an NVIDIA RTX 3090 (24\,GB), a 12-core AMD Ryzen 9 5900X and
128\,GB RAM, under Python 3.11 with CUDA 12.8, CuPy, RAPIDS 26.2 (cuGraph,
cuOpt), OptiX 8.0 via \code{triro}, HiGHS via \code{highspy}, OMPL 2.0 and
Isaac~Sim 5.0 for replay. Code, experiment scripts and the raw per-run JSON
behind every number below are on the \code{camera-ready/experiments} branch of
\url{https://github.com/krystofpsenicka/3D-Inspection}, with results under
\code{experiments/results/}.
Most experiments use the \emph{Duke of Lancaster} shipwreck (photogrammetric,
${\sim}584$k triangles, rescaled to a 50\,m longest axis), whose smooth hull and
topologically complex regions stress both visibility and routing; cross-model
experiments add the nine TOSCA meshes~\cite{bronstein2008} (10\,m) and four
dense laser scans of $0.35$--$7.2$M faces (\code{armadillo}, \code{dragon},
\code{happy\_buddha}, \code{xyz\_dragon}, 20\,m), used courtesy of the Stanford
Computer Graphics Laboratory~\cite{stanford3dscanrep}. Duke is our only real
underwater structure; the other thirteen meshes are not underwater and are not
meant as domain benchmarks. They vary two things the planner is sensitive to and
the wreck fixes --- surface topology (concavities, thin features) and
mesh resolution --- so they test whether the pipeline degrades on geometry
unlike its tuning set, not whether it works underwater. Unless stated, defaults are
$|P|=200{,}000$ / $|C|=1500$ on Duke ($50{,}000$ / $500$ elsewhere),
$\alpha=0.95$, $K=5$, $\beta=0.5$, \code{weighted\_curvature} sampling, GPU
raycasting, lazy-greedy set cover, and cuOpt with a 120\,s limit and 5\% gap
target. Constants: $r_{\text{robot}}=0.35$\,m, hence
$d_{\text{safe}}=0.70$\,m; voxels $0.10$\,m (fine) and $0.50$\,m (MAPF coarse)
with $\Delta_t=0.25$\,s; cruise $2$\,m/s; dwell $2$\,s; control rate
$\Delta_{\text{traj}}=0.02$\,s.

\begin{figure}[t]
\centering
\includegraphics[width=0.55\linewidth]{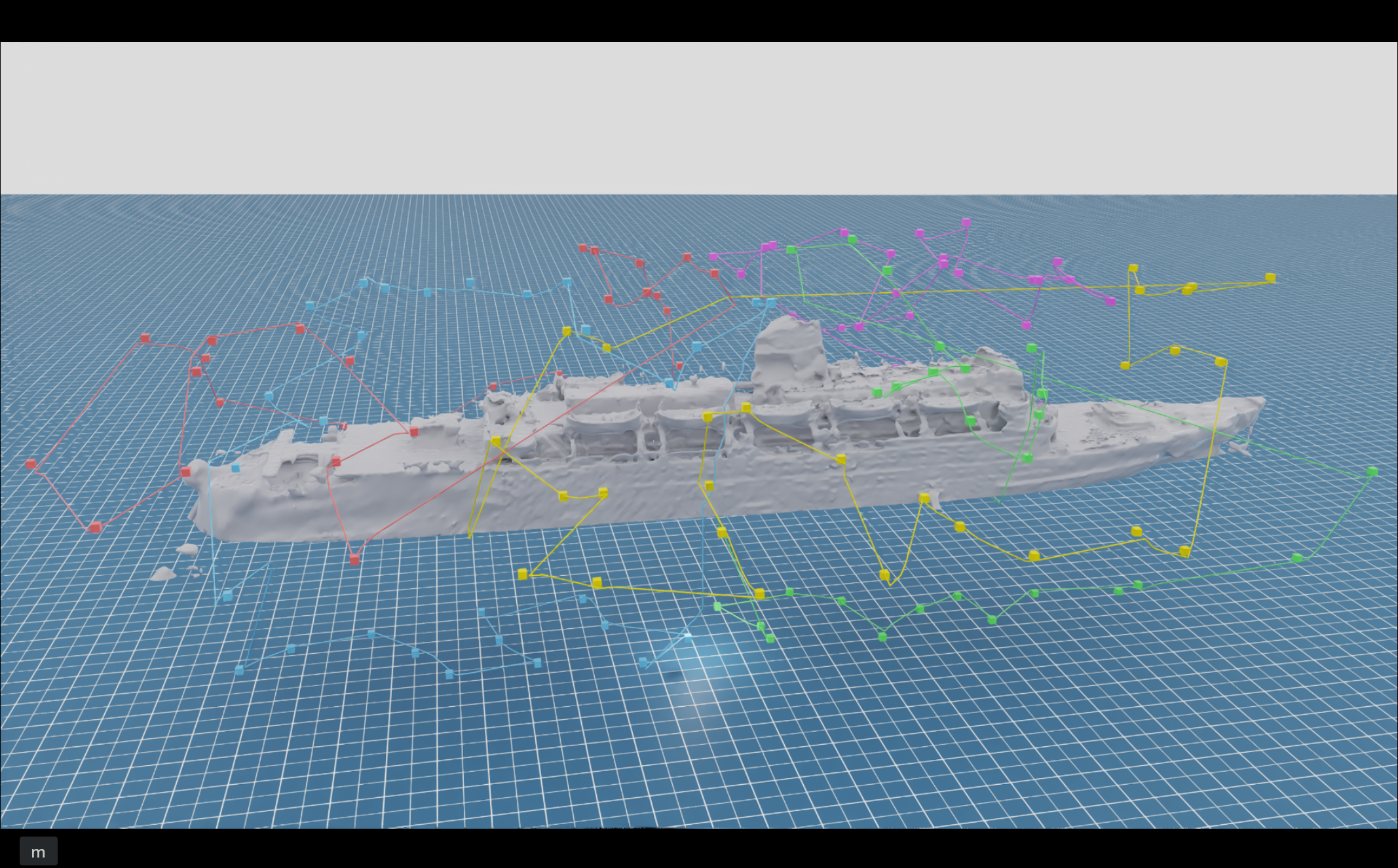}
\caption{Isaac~Sim preview of a $K{=}5$ Duke of Lancaster run. Coloured
polylines are the MAPF trajectories; vertices are inspection viewpoints.}
\label{fig:isaac}
\end{figure}

\paragraph{Units and statistics.}
Stage-4 (VRP) quantities are \emph{distances in metres} --- the routing model
never sees time --- while stage-5 (MAPF) and end-to-end quantities are
\emph{times in seconds}, from the executed trajectory at the cruise speed and
dwell above; where a quantity is called ``makespan'', the unit says which stage
it belongs to. All results are mean $\pm$ std over independent seeds, with $n$
stated. Seed budgets follow cost: 10 for the cheap sampling/visibility/set-cover
experiments (E01, E02, E05, E06), 5 for routing and scaling (E07, E08, E11, E14)
and 3--5 for the MAPF-bound ones (E10, E12, E13, E15, E17), one Duke run at 20
orderings costing ${\sim}21$ minutes. With $n\le10$ we make no significance
claims: we report the spread and say where configurations overlap within one
standard deviation. Figure~\ref{fig:isaac} shows a $K{=}5$ Duke run.

\paragraph{Sampling strategies (E01, 10 seeds).}
With coverage at or above target everywhere, the selected viewpoint count
measures efficiency directly. On Duke, curvature weighting drops it from
$227.4\pm11.7$ (unbiased \code{weighted}) to $210.1\pm4.3$ ($-7.6\%$) and CMA-ES
to $124.6\pm1.6$ ($-45\%$) --- but the latter costs $656\pm196$\,s against
$1.14\pm0.13$\,s (${\sim}570\times$), with visibility and set cover invariant.
We therefore default to \code{weighted\_curvature}, using CMA-ES only when a
${\sim}10$-minute sampling budget is acceptable.

\paragraph{Visibility (E05, 10 seeds).}
Raycasting agrees exactly with ground truth (Table~\ref{tab:e05}); both
$\varepsilon$-visibility variants drop to $0.875\pm0.005$ on the cluttered Duke
geometry versus $0.949\pm0.006$ on TOSCA. Their errors are false-positive
``visible'' labels, which make set cover terminate believing the target is met:
achieved coverage undershoots the $95\%$ target by $3.8\pm1.6$ percentage points
on Duke and $1.3\pm0.5$ on TOSCA, against $0.0$ for raycasting. GPU raycasting
is also $1.9\times$ faster than GPU $\varepsilon$-visibility and
$162$--$218\times$ faster than either CPU variant --- simultaneously the fastest
and the most accurate option whenever a mesh is available.
$\varepsilon$-visibility remains the fallback for bare point clouds.

\begin{table}[t]
\centering
\caption{E05: visibility models vs.\ exhaustive BVH ray--triangle ground truth;
mean $\pm$ std over 10 seeds $\times$ 5 coverage targets. \code{raycast} vs.\
\code{epsilon} are different \emph{models} (BVH occlusion tracing against the
mesh vs.\ Lien's approximation on the oriented point set); \code{gpu} vs.\
\code{cpu} are \emph{implementations} of each --- so $F_1$ gaps between models
are modelling error, and time gaps within a model pair are implementation
speedups. Times are the full $N\times M$ query on Duke; TOSCA pools
\code{cat0}, \code{david0}, \code{wolf0}.}
\label{tab:e05}
\footnotesize
\begin{tabular}{@{}lccc@{}}
\toprule
Model / impl. & $F_1$ Duke & $F_1$ TOSCA & Duke query time (s) \\
\midrule
\code{gpu\_raycast}  & $1.000\pm0.000$ & $1.000\pm0.000$ & $0.0196\pm0.0013$ \\
\code{cpu\_raycast}  & $1.000\pm0.000$ & $1.000\pm0.000$ & $3.166\pm0.103$ \\
\code{gpu\_epsilon}  & $0.875\pm0.005$ & $0.949\pm0.006$ & $0.0369\pm0.0052$ \\
\code{cpu\_epsilon}  & $0.875\pm0.019$ & $0.949\pm0.006$ & $4.259\pm0.129$ \\
\bottomrule
\end{tabular}
\end{table}

\paragraph{Set cover (E06, 10 seeds, $\alpha=0.95$).}
Greedy, GPU greedy and lazy-greedy return identical viewpoint counts on every
mesh ($209.6\pm10.7$ on Duke, $16.8$--$23.6$ on TOSCA): the lazy update changes
speed, not the optimum. The expansion variants cut the Duke count to
$147.3\pm2.8$ ($-30\%$) and $132.1\pm2.7$ ($-37\%$, CMA-ES re-sampling). Greedy's
integrality gap against the set-cover LP relaxation is $1.13\times$ on Duke and
$1.12$--$1.20\times$ on TOSCA, far inside the $\ln n$ guarantee. On Duke
lazy-greedy runs in $0.155\pm0.036$\,s --- $502\times$ faster than plain greedy
($77.7\pm50.5$\,s) and $5.9\times$ faster than the GPU scan ($0.905\pm0.159$\,s)
--- so the heap beats the dense GEMV at $|C|=1500$.

\begin{figure*}[t]
\centering
\begin{subfigure}[b]{0.32\linewidth}
\centering
\includegraphics[width=\linewidth]{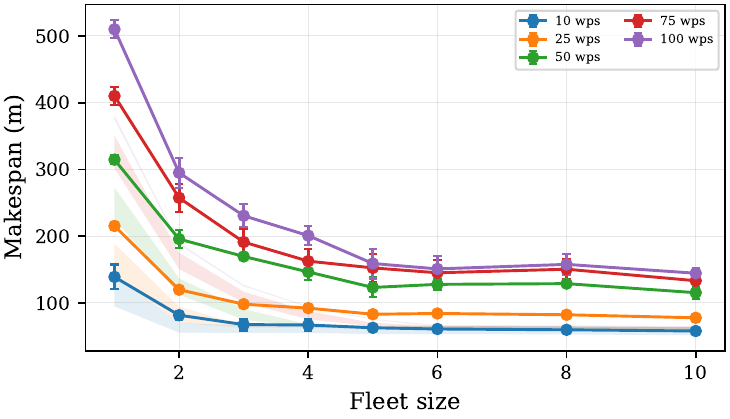}
\caption{E07: routing-stage makespan (longest single route, metres) vs.\ fleet
size $K$; the knee at $K=5$--$6$ sets the default. 5 seeds.}
\label{fig:e07}
\end{subfigure}
\hfill
\begin{subfigure}[b]{0.32\linewidth}
\centering
\includegraphics[width=\linewidth]{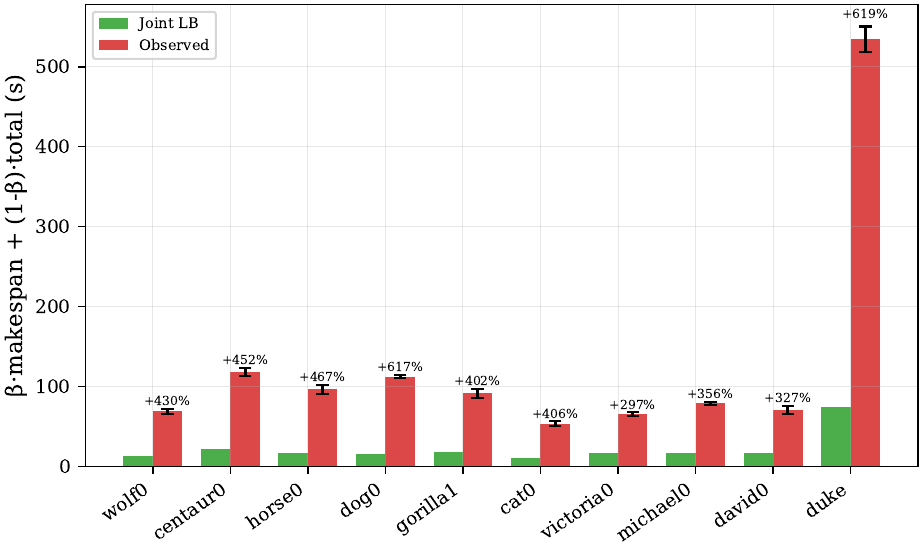}
\caption{E10: observed $\beta$-blended objective (s) vs.\ the analytical bound
$L^\star$ of Eq.~\eqref{eq:jointlb}, per mesh: a $4\times$--$7\times$ band.}
\label{fig:e10gap}
\end{subfigure}
\hfill
\begin{subfigure}[b]{0.32\linewidth}
\centering
\includegraphics[width=\linewidth]{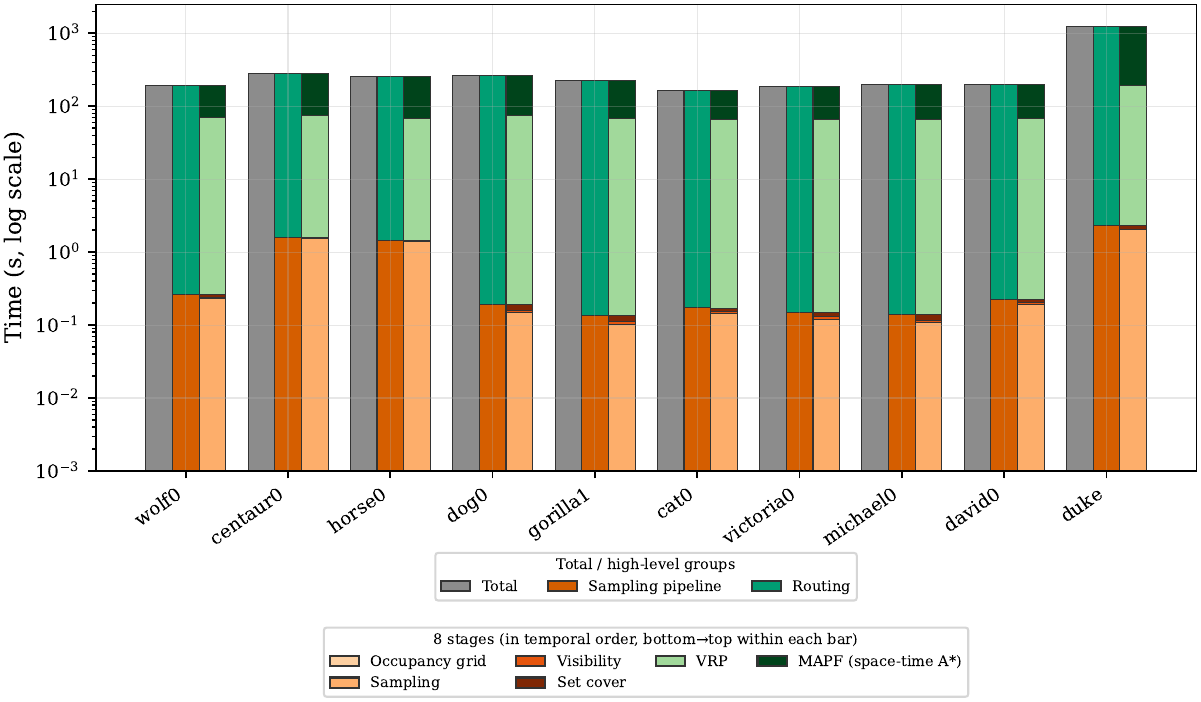}
\caption{E10: per-stage timing per mesh (log scale); MAPF dominates.}
\label{fig:e10}
\end{subfigure}
\caption{Fleet-size scaling (left), end-to-end gap against the analytical lower
bound (centre), per-stage timing (right).}
\label{fig:quality}
\end{figure*}

\paragraph{Fleet scaling and blend (E07/E08, 5 seeds).}
Quantities here are routing-stage distances. At 100 waypoints
(Figure~\ref{fig:e07}) makespan drops $69\%$ from $509.9\pm13.6$\,m at $K=1$ to
$159.0\pm22.4$\,m at $K=5$, then only $9\%$ more to $144.5\pm8.1$\,m at $K=10$;
the knee at $K=5$--$6$ motivates the default. Total distance grows
correspondingly ($509.9 \to 688.5\pm85.1 \to 1082.7\pm49.2$\,m). Distance-matrix
construction ($60$--$65$\,s) and the cuOpt solve (budget-bound at
${\sim}121$\,s) are constant in $K$; only per-robot trajectory planning is
$K$-dependent, growing from $24$\,s at $K=1$ to $1308$\,s at $K=10$ and
dominating from $K\ge4$. Sweeping $\beta$ over $[0,1]$ (E08) moves the mean
makespan non-monotonically between $121.3\pm9.0$ and $144.5\pm18.4$\,m and total
distance between $447\pm21$ and $594\pm88$\,m --- comparable to the seed-to-seed
spread at fixed $\beta$. The cause is the budget, not the blend: for all 13
values of $\beta$, all 59 successful runs stopped on the time limit rather than
the 5\% gap target ($120.85\pm0.25$\,s of 120\,s). Within this budget the solver
therefore returns whatever incumbent branch-and-bound has reached, and we report
no numeric optimality gap for the routing stage; E10 measures solution quality
against independent lower bounds instead. Two of 61 runs are excluded
as cuOpt CUDA failures.

\paragraph{Ablation of the routing cuts (E11/E15/E16, 5 seeds).}
Both cuts are listed as contributions, so we ablate them: the \emph{same} seeded
Duke sub-instance is solved four ways --- \code{none} (plain lifted MTZ),
\code{reach} (reachability filter only), \code{pair} (forbidden-pair cuts only),
\code{both} (default). At the 120\,s production budget and 50 waypoints the four
are \emph{indistinguishable} ($309.8\pm15.9$, $306.9\pm15.5$, $311.5\pm16.3$ and
$317.7\pm28.3$\,m, every mean inside one standard deviation of every other, all
40 runs exhausting the budget), and at 1800\,s (E15, 3 seeds) \code{both}
reaches $299.1\pm18.0$\,m against $307.1\pm9.2$\,m for \code{none} --- again
inside one standard deviation.

E16 explains why, by inspecting the \emph{models} rather than the solutions
(Table~\ref{tab:e16}). The $\beta$-aware filter drops \emph{zero} arcs at every
size from 10 to 50 waypoints: the warm-start bound
$T_{\text{tour}}^{\text{ub}}$ is never tight enough for
$c_{d_v,i}+c_{ij}+c_{j,d_v}$ to exceed it. The forbidden-pair cuts do fire, but
only on small instances, and the count collapses with size. At our production
size of 50 waypoints all four variants therefore build the \emph{identical}
model ($12{,}801$ variables, $12{,}615$ rows), so the spread above is solver
nondeterminism rather than a difference between formulations, and the root LP
bound is identical across configurations to solver tolerance at every size.
Both cuts are valid --- at 10 waypoints, where the MIP closes to proven
optimality, all four return a bit-identical optimum --- but on this problem they
are inert at the sizes that matter, because the bound they are derived from
never binds.

\paragraph{Ablation of the parallel-frontier search (E17, 3 seeds).}
E12 below isolates the continuous-time conflict filter; E17 isolates the search
itself. We intercept the single-agent queries of a real Duke $K{=}5$ plan and
solve each three ways on identical inputs: \code{gpu\_pf} (production,
$f_\Delta=2.0$), \code{gpu\_seq} (the same code with $f_\Delta=0$, one $f$-layer
at a time) and \code{cpu\_seq} (a sequential heap A* over the same state, cost
model, heuristic and filter). Duke's legs are small at the $0.50$\,m MAPF
resolution, so we report the sharper probe: 60 legs on an $8\times$ finer grid
($317\times113\times106$). The parallel frontier is decisive and free: \code{gpu\_seq}
takes $38.0\times$ longer at the median ($78.1\times$ mean, $664\times$ worst)
and degrades catastrophically on hard queries ($276$\,s worst leg against
$0.42$\,s), at a cost ratio of $1.0000$ on every leg --- so the
expansion strategy, not the port, makes the stage tractable, and the
$f_\Delta$-suboptimality of Section~\ref{sec:mapf} never binds. Against the CPU
the answer is narrower: the median leg \emph{favours} \code{cpu\_seq}
($0.46\times$) and the mean is parity ($0.92\times$), a $0.03$--$0.1$\,s launch
overhead dominating a millisecond search. What the GPU buys
is the tail --- its worst leg is bounded at $0.42$\,s against the CPU's
$3.2$\,s, and on the three hard legs (CPU ${>}0.5$\,s) it is $5.4\times$ faster.

\begin{table}[t]
\centering
\caption{E16: root LP relaxation per instance size $|W|$, 5 seeds $\times$ 4 cut
configurations (100 LP solves, all optimal). \emph{Arcs dropped} counts binary
variables removed by the $\beta$-aware filter; \emph{rows added} counts
forbidden-pair cuts. The LP bound is identical for all four configurations at
every size (relative spread $<10^{-8}$), so one column suffices: neither cut
tightens the relaxation, and at $|W|=50$ neither changes the model at all.}
\label{tab:e16}
\footnotesize
\begin{tabular}{@{}rrrc@{}}
\toprule
$|W|$ & arcs dropped (\code{reach}) & rows added (\code{pair}) & root LP bound (m) \\
\midrule
10 & 0 & 19  & $40.43\pm3.37$ \\
15 & 0 & 120 & $44.41\pm2.85$ \\
20 & 0 & 88  & $47.27\pm1.58$ \\
25 & 0 & 2   & $47.02\pm2.76$ \\
50 & 0 & 0   & $57.31\pm1.56$ \\
\bottomrule
\end{tabular}
\end{table}

\paragraph{Scaling in $|C|$ and $|P|$ (E14, 5 seeds).}
Sweeping one axis at a time on Duke, visibility and set-cover time are linear in
both $|C|$ ($t_{\text{vis}}\!:0.008\!\to\!0.055$\,s,
$t_{\text{opt}}\!:0.17\!\to\!0.63$\,s for $|C|=500\!\to\!4000$) and $|P|$
($0.007\!\to\!0.042$\,s, $0.12\!\to\!0.52$\,s for $|P|=50$k$\to\!400$k), as the
$O(NM)$ kernel and GEMV costs predict. Quality behaves differently along the two
axes. In $|P|$ the viewpoint count is essentially flat ($197.6\pm6.8$,
$202.8\pm7.4$, $211.6\pm10.0$, $220.2\pm5.8$ at $50$k$/100$k$/200$k$/400$k --- an
$11\%$ drift over an $8\times$ increase, coverage $0.950$ throughout). In $|C|$ the count keeps
falling ($366.6\pm12.1$, $293.2\pm47.9$, $215.2\pm16.3$, $178.6\pm2.4$,
$167.6\pm1.9$ at $500/1000/1500/2500/4000$) and $|C|=500$ misses the coverage
target outright ($0.895\pm0.007$). $|C|$ is thus a real quality lever that is
cheap upstream --- $1500\to4000$ costs $0.4$\,s and removes $22\%$ of the
viewpoints, shrinking the routing instance --- so our default of $1500$ is
conservative.

\paragraph{Sampler routing impact (E09).}
End to end on Duke, the probabilistic samplers cluster on the high-makespan side
of the valid frontier, while a travel-weight-tuned CMA-ES
(\code{travel\_weight}${=}0.02$) cuts makespan by $22\%$ while holding coverage
--- a useful lever, but brittle: $0.03$ collapses coverage to ${\sim}81\%$.

\paragraph{Coordination and post-smoothing safety (E12/E13).}
Collision freedom must be checked on the trajectories actually emitted --- after
smoothing and densification --- not on the coarse grid where the conflict test
runs. E12 plans each Duke instance twice at $K=2,\ldots,6$ (3 seeds):
\emph{coordinated} (each robot checked against its predecessors' committed
motion) and \emph{independent} (each planned as if alone). Coordinated plans
contain $0.00\pm0.00$ colliding pairs at every $K$, minimum separation
$0.73$--$1.28$\,m against $d_{\text{safe}}=0.70$\,m; independent plans contain
$0.67\pm0.58$, $1.00\pm1.00$, $1.00\pm1.00$, $2.33\pm0.58$, $2.67\pm0.58$ pairs
at $K=2,\ldots,6$. E13 then audits the final densified $K{=}5$ output of
the full 20-ordering pipeline (5 seeds): $0$ environment penetrations, $0$
colliding pairs, $0$ breaches of $d_{\text{safe}}$, minimum
separation $0.872\pm0.154$\,m (worst single seed $0.704$\,m), $0$ A* failures;
and of $215\pm11$ waypoints per mission $0$ are missed. Coverage enters only
through that last quantity: we certify it at the selected viewpoints and check
that smoothing still reaches each, rather than recomputing visibility along the
smoothed path. Smoothing after coordination is therefore safe here, because a
smoothed leg that would break separation is rejected for its A* polyline.

\paragraph{Priority-ordering budget (E12, orderings arm, 3 seeds).}
Sweeping the budget $n\in\{1,5,10,20\}$ on Duke shows the orderings are a
\emph{feasibility} lever, not a cost lever (Table~\ref{tab:e12ord}). The blended
objective is flat, a $0.2\%$ change across the whole sweep. What changes is
collision freedom: at $n{=}1$ two of three seeds finish with residual collisions
and a minimum separation below $d_{\text{safe}}$, whereas every $n\ge5$ run is
conflict-free with margin. The gain saturates immediately --- the $n{=}5$, $10$
and $20$ plans agree to three significant figures --- so, with MAPF wall-clock
linear in $n$, our default of 20 is four times more expensive than it needs to
be on these instances. $n{=}5$ is the better setting, and a single ordering is
not enough to rely on. Every other experiment here uses the shipped default of
20.

\begin{table}[t]
\centering
\caption{E12 orderings arm: priority-ordering budget $n$ on Duke, $K{=}5$, 3
seeds, measured on the final smoothed trajectories ($d_{\text{safe}}=0.70$\,m).
The objective is flat in $n$; collision freedom is not.}
\label{tab:e12ord}
\footnotesize
\begin{tabular}{@{}rcccc@{}}
\toprule
$n$ & objective (s) & makespan (s) & colliding pairs & min.\ separation (m) \\
\midrule
1  & $537.6\pm26.7$ & $190.3\pm16.9$ & $1.00\pm1.00$ & $0.52\pm0.28$ \\
5  & $536.7\pm26.3$ & $190.0\pm16.9$ & $0.00\pm0.00$ & $0.74\pm0.01$ \\
10 & $536.7\pm26.3$ & $190.0\pm16.9$ & $0.00\pm0.00$ & $0.76\pm0.04$ \\
20 & $536.6\pm26.2$ & $190.0\pm16.9$ & $0.00\pm0.00$ & $0.74\pm0.01$ \\
\bottomrule
\end{tabular}
\end{table}

\paragraph{Cross-model generalization (E10).}
We run the full pipeline on 14 meshes --- Duke, nine TOSCA meshes and four dense
laser scans --- and compare the observed $\beta$-blended objective (in seconds)
against a composite analytical lower bound of three independently valid terms.
Let $N_{\text{lb}}$ bound the pose count from below, the larger of an
information-theoretic count $\lceil \alpha M / \max_j |V_j| \rceil$ and a
partial-cover antichain bound $\max(0,|S| - \lceil(1-\alpha)M\rceil)$ with $S$ a
greedily built set of target points no single candidate sees two of; let
$a(\mathbf{p}) = \max(0, \min_v \lVert \mathbf{x}_v^0 - \mathbf{p}\rVert - z_f)$
be the unavoidable approach distance from the nearest depot given the far plane
$z_f$, and $a^\star$ the $\lceil\alpha M\rceil$-th smallest such value (an
adversary may skip the $\lceil(1-\alpha)M\rceil$ hardest targets). With cruise
speed $s$ and dwell $\tau$,
\begin{equation}\label{eq:jointlb}
\begin{gathered}
L_{\text{tot}} = N_{\text{lb}}\tau + 2a^\star\!/s, \quad
L^\star = \beta L_{\text{mks}} + (1-\beta) L_{\text{tot}},\\
L_{\text{mks}} = \max\!\bigl(N_{\text{lb}}\tau/K,\; 2a^\star\!/s + \tau,\;
                  L_{\text{tot}}/K\bigr).
\end{gathered}
\end{equation}
Against $L^\star$ the observed objective sits in a $4.0\times$--$7.2\times$ band
across the 14 meshes (Figure~\ref{fig:e10gap}). That width is attributable, and
it is not a routing or path-finding failure: against a \emph{tighter} bound
reusing the selected viewpoints and true voxel-graph distances (the 2-factor LP
relaxation of the multi-depot VRP, mapped to seconds at $s$ and $\tau$), the
same plans sit in a much narrower $1.26\times$--$1.57\times$ band. What varies
is the pose-count term: the ratio of selected viewpoints to $N_{\text{lb}}$
ranges from $2.43$ (\code{cat0}) to $4.59$ (\code{dog0}) and correlates with the
joint gap at $r=0.76$. The band therefore measures how loose the antichain pose
bound is on a given geometry --- greedy set cover is within $1.13\times$ of the
LP optimum (E06) --- not how far the plan is from optimal.

Viewpoint count tracks physical surface area, not face count: Duke ($584$k
faces, $50$\,m) needs $207.6\pm6.9$ viewpoints, \code{xyz\_dragon} ($7.2$M
faces, $20$\,m) only $79.0\pm3.0$, \code{cat0} ($56$k, $10$\,m) $17.0\pm0.8$.
Mesh resolution is thus almost free --- a $12\times$ denser mesh than Duke costs
$5.2\pm7.2$\,s of sampling/visibility/set-cover --- which is what one wants from
a pipeline fed by photogrammetry. MAPF dominates wall-clock time ($84.2\pm1.4\%$ on
Duke, $59$--$79\%$ elsewhere), VRP second (Figure~\ref{fig:e10}); the sampling
group sums to $2.3\pm1.3$\,s, so further effort should target MAPF and VRP.

\section{Conclusion}
\label{sec:conclusion}

We presented \textsc{3D-Inspection}, a GPU-resident pipeline turning a triangle
mesh and a fleet size into 6-DOF inspection trajectories by decomposing the
NP-hard joint problem into five stages. Our algorithmic contributions are a GPU
batch adaptation of Lien's $\varepsilon$-visibility, a $\beta$-aware per-vehicle
tour bound with forbidden-pair cuts for min-max VRP, and a 4D space-time port of
Zhou and Zeng's parallel-frontier A* with continuous-time conflict checking.
Across 14 meshes: curvature-weighted sampling is a robust default; GPU
raycasting is simultaneously the fastest and most accurate visibility model;
lazy-greedy lands within $1.13\times$ of the LP relaxation; routing makespan
saturates near $K=5$; the coordination layer removes every inter-robot collision
independent planning would leave, and the audited post-smoothing output has zero
collisions and zero missed waypoints; the parallel frontier is worth a median
$38\times$ at zero cost in path quality, though against a CPU A* it wins on
worst-case latency, not median speed; and the end-to-end gap against an
analytical joint bound stays in a $4\times$--$7\times$ band whose width is
attributable to the pose-count term of that bound, with MAPF dominating
wall-clock time. The two routing cuts, however, turn out to be inert at
production instance sizes: the warm-start bound they derive from never binds.

\paragraph{Limitations.}
The formulation is open-loop and geometry-only, so the assumptions of
Section~\ref{sec:intro} bound what these results mean; three would break first
in a real mission. \emph{(i)~Localization and map error:} coverage is certified
against the prior mesh, so a pose error comparable to the $6$\,m far plane or a
mis-registered map invalidates both the coverage guarantee and the $0.70$\,m
separation margin, and nothing in the pipeline detects this --- an
active-perception layer that revises the plan against the map being
built~\cite{englot2013,hernandez2015} is the natural next layer above ours.
\emph{(ii)~Disturbances:} currents, thruster saturation and turbidity are absent
from both model and evaluation; margins are constants, not optimized. \emph{(iii)~Vehicle class:} we assume untethered AUVs, whereas for the tethered
ROVs common in practice tether length, snag avoidance and operator workflow
would dominate and reshape the routing formulation. Beyond modelling, the
decomposition costs global optimality, priority-based MAPF is incomplete and
would need a CBS variant for denser fleets, and most stages are slow or
infeasible on CPU-only hardware.

We also compare only against lower bounds and variants of our own pipeline, not
an external planner. Published structural-inspection planners
(e.g.~\cite{englot2013,bircher2015}) are single-robot, optimize coverage-only or
next-best-view objectives rather than a min-max fleet objective, and are
evaluated on their own vehicle and sensor models, so a head-to-head number would
mostly measure our reimplementation of them. Lower bounds avoid that but only
certify a factor from optimal; showing we advance the state of the art would
need a shared multi-robot 3D-inspection benchmark with a common sensor model and
objective, which does not yet exist.

\paragraph{Future work.}
A location-routing formulation coupling viewpoint selection with travel cost; a
CBS-based MAPF for dense fleets; a longer-horizon or decomposition-based VRP
solve, since our routing results are budget- rather than formulation-bound;
an active-perception layer and currents-aware planning toward deployment; and
differentiable visibility for gradient-based pose refinement.

\begin{acknowledgments}
We thank J\'achym Herynek for valuable technical discussions.
\end{acknowledgments}

\section*{Declaration on Generative AI}
During the preparation of this work, the authors used Claude (Anthropic) in
order to: drafting content, paraphrase and reword, improve writing style,
formatting assistance, content analysis, and code generation. After
using this tool, the authors reviewed and edited the content as needed and take
full responsibility for the publication's content.

\bibliography{bibliography}

\end{document}